\pdfoutput=1
\documentclass[11pt]{article}

\usepackage[final]{acl}
\usepackage{graphicx} 

\usepackage{times}
\usepackage{latexsym}
\usepackage[normalem]{ulem}
\useunder{\uline}{\ul}{}
\usepackage[T1]{fontenc}
\usepackage[utf8]{inputenc}

\usepackage{microtype}

\usepackage{inconsolata}
\usepackage{amsmath}

\usepackage{amsfonts}
\usepackage{algorithm}

\usepackage{algpseudocode}

\title{Selective Prompting Tuning for Personalized Conversations with LLMs}

\author{Qiushi Huang\textsuperscript{\rm 1,2}, Xubo Liu\textsuperscript{\rm 1}, Tom Ko\textsuperscript{\rm 3}, Bo Wu\textsuperscript{\rm 4}, Wenwu Wang\textsuperscript{\rm 1}, \\ \bf  Yu Zhang\textsuperscript{\rm 2}\thanks{\ \ Corresponding authors.}  , Lilian Tang\textsuperscript{\rm 1$\ast$  } \\
\textsuperscript{\rm 1}University of Surrey,
    \textsuperscript{\rm 2}Southern University of Science and Technology,\\
    \textsuperscript{\rm 3}ByteDance AI Lab,
    \textsuperscript{\rm 4}MIT-IBM Watson AI Lab \\
    \{qiushi.huang, xubo.liu, w.wang, h.tang\}@surrey.ac.uk, \\\{tomkocse, yu.zhang.ust\}@gmail.com, bo.wu@ibm.com
    }

\begin{document}
\maketitle
\begin{abstract}
% REVISION NEEDED AFTER REVISING INTRODUCTION. 

In conversational AI, personalizing dialogues with persona profiles and contextual understanding is essential. Despite large language models' (LLMs) improved response coherence, effective persona integration remains a challenge. In this work, we first study two common approaches for personalizing LLMs: textual prompting and direct fine-tuning. We observed that textual prompting often struggles to yield responses that are similar to the ground truths in datasets, while direct fine-tuning tends to produce repetitive or overly generic replies. To alleviate those issues, we propose \textbf{S}elective \textbf{P}rompt \textbf{T}uning (SPT), which softly prompts LLMs for personalized conversations in a selective way. Concretely, SPT initializes a set of soft prompts and uses a trainable dense retriever to adaptively select suitable soft prompts for LLMs according to different input contexts, where the prompt retriever is dynamically updated through feedback from the LLMs. Additionally, we propose context-prompt contrastive learning and prompt fusion learning to encourage the SPT to enhance the diversity of personalized conversations. Experiments on the CONVAI2 dataset demonstrate that SPT significantly enhances response diversity by up to 90\%, along with improvements in other critical performance indicators. Those results highlight the efficacy of SPT in fostering engaging and personalized dialogue generation. The SPT model code is publicly available for further exploration. \footnote{\url{https://github.com/hqsiswiliam/SPT}}

\end{abstract}

% https://math.stackexchange.com/questions/877802/rank-of-a-matrix-sum

\section{Introduction}

% briefly introduce LLM & the potential applications of the personalized dialogue generation
% Recent advancements in large language models (LLMs) have markedly improved dialogue generation, leading to a significant shift in their ability to engage in coherent and interactive conversations. However, despite these advancements, the integration of personalization within LLMs remains largely unexplored, particularly in their predominant role as tools for query resolution. This gap underscores the untapped potential for developing personalized dialogue systems. Personalization is essential in diverse applications, from creating intelligent non-player characters in gaming environments, who can respond with persona-specific dialogues, to serving as empathetic companions for the elderly, potentially improving their mental health. Furthermore, personalized interactions are key to humanizing user experiences in both entertainment and healthcare platforms, enhancing engagingness and effectiveness.

% Introduction to personalization
Personalization in dialogue systems enhances user interaction by creating a coherent and customized experience. It involves adapting conversations to individual preferences, backgrounds, and real-time context, ensuring each dialogue feels personally relevant. This tailored approach fosters a deeper connection between users and technology, making interactions more intuitive and engaging. By understanding and anticipating user needs, personalized dialogues can offer more than just relevant responses; they provide a seamless, conversational experience that mirrors human interaction, enriching the overall quality of digital communication.

% Introduction to personachat
PersonaChat \cite{PersonaChat} has become a pivotal dataset for personalization research in conversational AI, offering persona profiles that detail an interlocutor's preferences and background in four to five sentences. These profiles guide conversational agents in creating dialogues that are both engaging and consistent with the persona's characteristics and prior conversational context. This area has seen diverse approaches for enhancing personalization, such as attention mechanisms \cite{PAA}, reinforcement learning with multiple rewards \cite{bert_over_bert, mutual_persona}, and persona profile enrichment through stories \cite{lapdog}, demonstrating the breadth of innovation in making interactions more personalized and meaningful.

% Using LLM to do personachat
% I need to think about putting the introduction to soft prompt tuning somewhere.
Recently, the advent of large language models (LLMs) \cite{opt,llama2} has opened new avenues for dialogue generation, offering the potential for creating conversations that align with human preferences. However, fully leveraging LLMs to achieve the level of personalization showed in PersonaChat is a promising yet underexplored area. Currently, LLMs are primarily guided by direct textual prompts or through parameter-efficient fine-tuning like prompt tuning \cite{prompt-tuning} that only tunes a few virtual tokens instead of whole LLMs for specific tasks. %suggesting a vast potential for innovation in personalizing conversational AI.
% Challenges for using LLM to train personachat
% First challenge is tackled by selective prompt tuning

However, designing personalized conversational agents with LLMs faces two main challenges. The primary issue lies in diverse settings in conversations, 
% marked by unique personas and conversation lengths. 
% The first one is the complexity of datasets
%\footnote{***it is better to change to `the complexity of the persona-based dialogue generation task': What about changing to "diverse dialogues"?} 
% like PersonaChat, 
which encompass a wide array of dialogues, each characterized by unique persona profiles and varying lengths of conversation. This diversity necessitates an understanding of the distinct conversational settings within the data. 
% and inherent patterns
%\footnote{***`pattern' seems unclear. what does it represent?: What about revised to conversational settings?} 
% within the data.
Through textual prompting, it is hard to guide the model to generate desired responses aligned with the target texts. Simply fine-tuning LLMs through prompt tuning without careful conversational setting
% \footnote{***pattern again} 
analysis risks producing responses that lack specificity and depth, resulting in a generic and bland generation.

% second is through context-prompt contrastive, avoid bland
Secondly, another equally critical challenge arises from the limitations inherent to the datasets used for persona-based dialogue generation. Typically small and lacking in diversity, these datasets can restrict the model's exposure to a wide range of conversational scenarios. When LLMs (e.g., Llama2-7B \cite{llama2}) are tuned through trainable soft prompts on PersonaChat, they risk overfitting to specific persona profiles. This overfitting manifests in the model's responses, which become repetitive and overly aligned with the persona, often at the cost of dynamic and contextually appropriate interactions. Although this might lead to improvements in metrics such as F1 or BLEU scores, it detracts from the overall diversity and engagingness of the dialogues, undermining the model's ability to emulate authentic human conversation.
% How our approaches tackle the challenges

To handle those two challenges when designing personalized conversations with LLMs, we propose a Selective Prompt Tuning (SPT) model.
Specifically, to tackle the first challenge, it is crucial to identify inherent data patterns without explicit annotations. % on which patterns are similar. 
To achieve this, it is intuitive to utilize a group of multiple soft prompts to handle different conversational settings when tuning the model in a parameter-efficient way. However, as previously mentioned, the annotations for the dialogue settings are missing and even hard to discover and annotate. %tuning multiple soft prompts would seem to be infeasible in such circumstances. 
If we naively concurrently tune all prompts without clear distinctions, this would yield only marginal differences compared with tuning one soft prompt. Therefore, to build effective multiple prompts to discover the inherent data pattern inside the personalized dialogue, the proposed SPT model utilizes a dense retriever to adaptively select a proper soft prompt from the soft prompt group based on the given input context. To distinguish the effectiveness of soft prompts, we utilize the loss from LLMs as feedback to guide the update of the dense retriever without explicit annotations. Based on this, the proposed SPT model could discover patterns intrinsically associated with different dialogues. %, allowing the model to adapt without predefined labels. 
In this way, the retriever and soft prompt group evolve together, benefiting from continuous interactions that enrich their capability to discriminate and generate diverse, contextually relevant responses. %During inference, this collaborative mechanism effectively chooses a good soft prompt, ensuring rich and engaging dialogue outcomes. 

To address the second challenge that LLM may overfit small-scale datasets such as PersonaChat, the proposed SPT method integrates two complementary mechanisms: context-prompt contrastive learning and prompt fusion learning. The context-prompt contrastive learning mechanism ensures diversity by encouraging the use of different soft prompts for varied dialogue contexts, preventing repetitive responses. Concurrently, prompt fusion learning aggregates all prompt predictions during back-propagation, optimizing towards a unified output. This dual strategy not only preserves response diversity across contexts but also enhances overall model performance, demonstrating their cooperative effectiveness in tackling overfitting while maintaining the performance.

By integrating the above two parts into the SPT method, experiments on the CONVAI2 dataset \cite{convai2} with LLMs %a cornerstone in personalized dialogue generation, are conducted using 
(i.e., Llama2 \cite{llama2} and OPT \cite{opt}) %to assess the impact of selective prompt tuning. The results 
not only demonstrate marked improvements in response diversity and engagingness but also indicate enhancements in other key performance metrics. Quantitatively, the proposed SPT model consistently outperforms baselines across models with various sizes. Moreover, SPT offers profound insights into different dialogue scenarios, particularly in the model's strategic prompt selection. Comprehensive ablation studies highlight the adaptability of different prompts to specific dialogue contexts. 

% Furthermore, the proposed SPT method effectively promotes diversity along with improvement in other critical metrics 

% and exploratory learning throughout the training phase.
%\footnote{***what does it mean?: Plan to add explore diverse soft prompt selections here. A bit twisted here, so Deleted.} 

Overall, our contributions can be summarized as follows.
\begin{itemize}

\item We present the novel SPT method 
% for personalized dialogue generation 
by integrating a trainable dense retriever with dynamic soft prompt selection to improve dialogue personalization and enhance both the diversity and engagingness.

\item In the proposed SPT method, we introduce the context-prompt contrastive mechanism and prompt fusion learning within a unified framework to foster prompt diversity and adaptability. %This innovative structure encourages the model to explore diverse soft prompt options when keeping the robust performance enhancement.

\item Extensive experiments show the effectiveness of the proposed SPT method.%Through extensive experiments on the CONVAI2 dataset with Llama2 and OPT models, we demonstrate our method's superiority over baselines in enhancing dialogue diversity, engagingness, and other crucial performance metrics. The integration of selective prompt tuning, prompt-contrasting, and prompt fusion learning mechanisms further exemplifies our approach's robustness in promoting comprehensive learning and diversity throughout the model's training phase.

\end{itemize}

\section{Related Work}
\subsection{Personalized Dialogue Generation}
The CONVAI2 dataset, curated from the PersonaChat dataset \cite{PersonaChat}, features a persona profile with four to five sentences for each interlocutor \cite{convai2}. This dataset has been established as a benchmark for personalized dialogue generation. Building upon this dataset, a variety of studies have explored different methods. For example, \citet{transfertransfo} extend the GPT2 model \cite{GPT2} with fine-tuning techniques specific to persona-based conversations. Differently, \citet{bert_over_bert} employed a tripartite BERT architecture \cite{BERT}, optimized through reinforcement learning, to craft responses. Similarly, \citet{mutual_persona} introduces a transmitter-receiver model by applying reinforcement learning with custom rewards to refine the dialogue generation process. \citet{d3} leverage model-agnostic data augmentation techniques to enrich the training dataset with pseudo-samples using models like GPT2 and BERT. \citet{PAA} develop an adaptive attention mechanism that coherently integrates persona and context information. \citet{lapdog} propose a LAPDOG method to incorporate an external story corpus to enhance persona profiles for richer response generation. In contrast to those methods, the proposed SPT framework decomposes the task with multiple soft prompts without necessitating additional annotations or reliance on external corpora, which enables the generation of diverse and engaging responses while maintaining the integrity of the conversational context. %\footnote {***what does `deconstruct' mean? need to be revised: What about " decomposes the task with multiple soft prompts without necessitating additional annotations or reliance on external corpora"}

\subsection{Language Models and Personalization}

Language models (LMs) estimate text sequence probabilities, with recent models expanding from millions \cite{GPT2, opt} to billions of parameters \cite{GPT3, opt}, and training corpora now including vast web texts and instructional data \cite{GPT3.5, llama2}. Such advancements have notably improved the performance of LMs on various NLP tasks, especially in generating high-quality text for conversational applications. While those LMs are adept at providing user-centric responses, personalization remains a challenge. The prevalent strategy involves appending manually crafted hard prompts to LMs, which is overly simplistic and can result in the `lost in the middle' problem \cite{lost_in_the_middle}. This occurs when the output of the LM is generically correct but lacks personalized context, struggling to reconcile broad training data with specific user prompts. 
To counteract this, the proposed SPT method enables the LLM to adapt its responses to varying personalized contexts more effectively. As a result, SPT fosters the generation of dialogue responses that are not only consistent but also highly personalized, addressing the core challenge of maintaining context relevance in user interactions.

\begin{figure*}[!tbph]
    \centering
    \includegraphics[width=\textwidth]{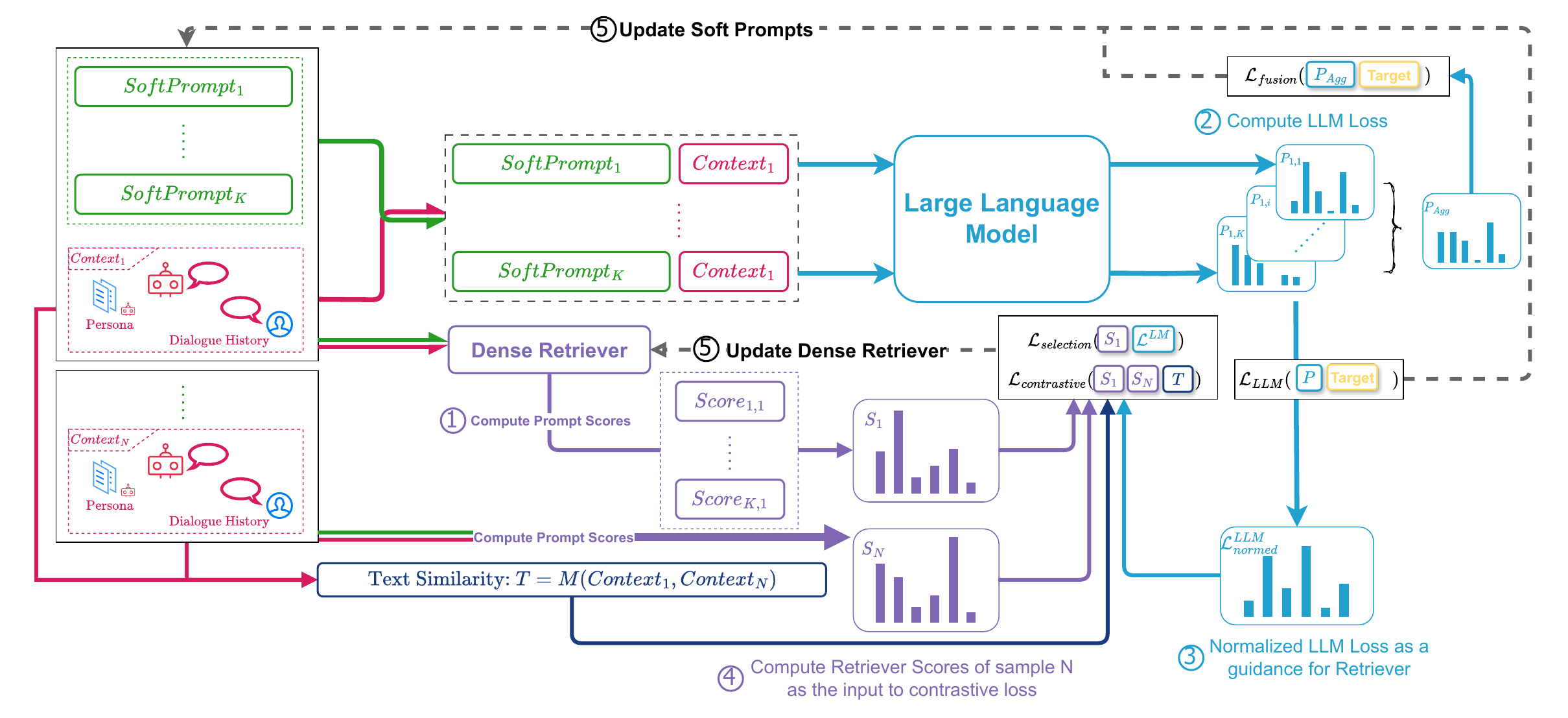}
    \caption{
    % An illustration of the proposed SPT method. %demonstration of training procedure of the selective prompt tuning on personalized dialogue generation task.
    Selective Prompt Tuning (SPT) process for personalized dialogue generation with large language models (LLMs). The process starts by computing similarity scores for $K$ soft prompts given the context, followed by LLM loss computation. The prompts are then fed into the LLM along with the context to generate multiple LLM losses which are normalized. A dense retriever computes another set of scores for a different context to inform the contrastive loss. The four computed losses guide the updates to the soft prompts and the retriever to enhance response diversity and relevance.
    }
    \label{fig:framework}
    \vspace{-5mm}
\end{figure*}

\section{Methodology}

In this section, we introduce the proposed SPT method.

\subsection{Problem Settings}

In persona-based dialogue sessions, a context is represented as $C=\{P,U\}$, where $P=\{p_1,\ldots,p_e\}$ denotes the persona comprising $e$ sentences (e.g., $4\le e\le 5$) to provide background information for a machine interlocutor $m$ and $U=\{u_{h,1}, u_{m,1},\ldots,u_{h,n}\}$ denotes the dialogue context initiated by the human $h$ to capture the exchange between human $h$ and machine $m$. The goal is to generate a machine's response $r=u_{m,n}$ that aligns with its persona $P$ and the context $U$. 

\subsection{Architecture}
Figure~\ref{fig:framework} illustrates the SPT framework, consisting of a soft prompt group, a dense retriever, and a frozen LLM. Within this framework, the dense retriever selects an appropriate soft prompt from the soft prompt group by determining the closest match to the given context $C$. The chosen prompt is then merged with $C$ to guide the LLM to produce compelling responses. 
The SPT framework restricts the soft prompt group and dense retriever to be trainable, while maintaining the LLM in a frozen state, which could significantly reduce the memory footprint and optimize resource utilization during training.

\paragraph{Soft Prompt Group} The soft prompt group, denoted by $SP=\{sp_1,...,sp_K\}$, consists of $K$ soft prompts with random initialization. Each prompt features $L \times D$ virtual tokens, where $D$ denotes the hidden dimension of the LLM and $L$ denotes the length of prompts. These prompts are fine-tuned during training while the LLM remains frozen.

\paragraph{Soft Prompt Selection} The soft prompt selection is done by a trainable retriever, $Ret(\cdot, \cdot)$, which calculates the similarity score $s_{C,sp}=\{s_{C,1}, ..., s_{C,K}\}$ between the context embedding $emb_C$ from the LLM and each candidate $sp_i$ in the soft prompt group $SP$. It ranks all the soft prompts based on the computed similarity score $\{s_{C,i}\}_{i=1}^K$ to identify the most suitable prompt for the context.

\paragraph{LLMs} The LLMs deployed here are the decoder-only causal language model with frozen weights and initialized from pre-trained models.
% Inputs, including the context and selected soft prompt, are concatenated and then fed to the LLM to generate the output.

\subsection{Computing Similarity between Soft Prompts and Context}
\label{sec:compute_sim}
To reduce computational overhead, the dense retriever $Ret$ utilizes two linear layers, i.e., $\mathrm{lin}_C$ and $\mathrm{lin}_{sp}$, for computing the similarity scores $\{s_{C,i}\}$. Those similarity scores are calculated using the context embedding $emb_C \in \mathbb{R}^{M\times D}$ obtained by the LLM's word embedding layer $\mathrm{LLM}_{emb}$ and the soft prompt representation in $\mathbb{R}^{L\times D}$. The similarity score is computed as
\begin{equation}
\begin{aligned}
emb_C &= \mathrm{LLM}_{emb}(C), \\
v_C &= \mathrm{lin}_C(emb_C), \\
v_{sp,i} &= \mathrm{lin}_{sp}(sp_i), \\
    \bar{v}_C &= \text{Avg}_{\text{dim}=0}(v_C), \\
    \bar{v}_{sp,i} &= \text{Avg}_{\text{dim}=0}(v_{sp,i}), \\
    s_{c,i}^{raw} &= \frac{\bar{v}_C \cdot \bar{v}_{sp,i}}{\|\bar{v}_C\|_2 \cdot \|\bar{v}_{sp,i}\|_2}, \\
    s_{C,i} &= \text{Softplus}(s_{C,i}^{raw}),
\end{aligned}
\end{equation}
where $\text{Avg}_{\text{dim}=0}(\cdot)$ denote the averaging operation along the length dimension to address the sequence length discrepancy between $emb_C$ and $sp_i$, $\text{Softplus}(\cdot)$ denotes the softplus activation function to ensure that $s_{C,i}$ remains in the range $[0, 1]$ and enhance the numerical stability during training, and $s_{C,i}$ represents the normalized similarity score between the context $C$ and the soft prompt $sp_i$. %yielding a consistent representation. 

\subsection{Learning Prompt Selection}

Navigating the lack of explicit annotations in complex dialogue scenarios poses a challenge in accurately guiding the retriever to assess the similarity between the context and each soft prompt. A naive method, which independently fine-tunes the entire soft prompt group and then selects candidates based on the similarity score during decoding, might lead to sub-optimal performance, akin to tuning a single soft prompt. To address this, we leverage context-driven losses from soft prompts, refining similarity score computations and enabling informed retriever decisions during training, as introduced in the next two subsections.

\subsubsection{Soft Prompt Loss}

For simplicity, consider the case with a single context. Given a context $c_n$ from persona and dialogue history and its corresponding ground truth response $target_n$, we calculate the negative log-likelihood loss for each soft prompt as
\begin{align}
pred_{i,n} &= \text{LLM}(\text{concat}(sp_i, c_n)),\nonumber\\
\mathcal{L}^{LLM}_i &= \text{NLL}(pred_{i,n}, target_n),\label{soft_prompt_loss}%\\
%    & \text{for each } i \in [1, K].
\end{align}
where $\text{concat}(\cdot,\cdot)$ denotes the concatenation operation, $\mathrm{LLM}(\cdot)$ denotes the LLM's forward operation, which takes a text sequence as the input and returns the predicted token probability distribution as the output, and $\text{NLL}(\cdot, \cdot)$ denotes the negative log-likelihood loss. %the LLM's forward operation
This process generates $K$ losses $\mathcal{L}^{LLM} = \{\mathcal{L}^{LLM}_1,...,\mathcal{L}^{LLM}_K\}$ to measure the predictive ability of each soft prompt.%correlating each soft prompt $sp_i$ with $c_n$ and $target_n$.

\subsubsection{Prompt Selection Loss}
In the absence of explicit annotations for conversational settings, updating the retriever to identify the most effective soft prompt for a given context is challenging. However, by using soft prompts in LLMs with the same context, the loss from different prompts can serve as a guide to determine which soft prompt is most suitable.
Based on this consideration, we use the soft prompt loss (i.e., $\mathcal{L}^{LLM}$ defined in Eq.~\eqref{soft_prompt_loss}) to gauge each candidate $sp_i$ in the soft prompt group $SP$ within $c_n$. Aligning the LLM's performance evaluation with the retriever's similarity scores is achieved by using the KL divergence between the negative language model loss (as guidance) and similarity scores. By denoting by $S_{c_n, SP}=[S_{c_n,sp_1},\ldots,S_{c_n,sp_K}]$ the similarity scores between $c_n$ and each $sp_i$ in $SP$, the prompt selection loss is formulated as
\begin{equation}
\begin{aligned}
\mathcal{L}^{LLM}_{normed}= \text{Softmax}(-\mathcal{L}^{LLM}/\tau_g),\\
    \mathcal{L}_{selection} = \text{KL}(S_{c_n, SP},\mathcal{L}^{LLM}_{normed}),
\end{aligned}
\end{equation}
where $\text{Softmax}(\cdot)$ denotes the softmax function, $\tau_g$ is a temperature hyper-parameter, and $\text{KL}(\cdot,\cdot)$ denotes the KL divergence.%controlling the sensitivity of the normalized language model loss.$\mathcal{L}^{LLM}_{normed}$ is the softmaxed negative language model loss, guiding retrieval similarity strength.  $\text{KL}$ represents the KL divergence loss, delineating the discrepancy between similarity score distributions and normalized language model losses. 
This loss is pivotal in ensuring the selections of the dense retriever are informed and coherent with the LLM, effectively mirroring the performance of soft prompts in generating contextually relevant and engaging responses.

\subsection{Context-Prompt Contrastive Learning}

While the aforementioned losses aid in training, there is a risk that the retriever often retrieves a single prompt and stagnates in such sub-optimal states. To alleviate this and foster prompt diversity to retrieve more prompts, we propose a context-prompt contrastive loss. This loss refines prompt selection by adjusting similarity scores based on the textual similarity of distinct contexts, thereby preventing to always select a single soft prompt and promoting varied selections. Specifically, the context-prompt contrastive loss dynamically recalibrates the similarity scores between pairs of context contents, %\footnote{***based on the definition, it is related to context but not prompts} 
considering their textual resemblance. %we distinguish between context pairs as either similar or dissimilar. , 
Mathematically, the context-prompt contrastive loss is formulated as
\begin{equation}
\small
\begin{aligned}
    \mathcal{L}_{con}(s_{c_i}, s_{c_j}) = \begin{cases}
    1 - \cos(s_{c_i}, s_{c_j}) & \text{if } M(c_i,c_j) > \Gamma \\ 
    \max(0, \cos(s_{c_i}, s_{c_j})) & \text{otherwise}
    \end{cases}
\end{aligned}
\label{context_CL_loss}
\end{equation}
where $M(\cdot,\cdot)$ denotes a distance function such as BLEU \cite{bleu}, $\Gamma$ denotes a threshold, $s_{c_i}$ denotes a vector of cosine similarity scores between a context $c_i$ and soft prompts in the soft prompt group, and $\mathrm{cos}(\cdot,\cdot)$ denotes the cosine similarity.

The function $\mathcal{L}_{con}$ amplifies the cosine similarity for similar context pairs (i.e., $M(c_i,c_j) > \Gamma$) and dampens it for dissimilar pairs (i.e., $M(c_i,c_j) \leq \Gamma$). This contrastive strategy not only ensures the retriever's alignment with the LLM's evaluations but also fosters a rich diversity and distinctiveness among different dialogue contexts, significantly bolstering the framework's overall adaptability.

\subsection{Prompt Fusion Learning}
To optimize the effectiveness of the soft prompts, we introduce a prompt fusion learning loss. This loss averages the predictive probabilities from all the soft prompts in the soft prompt group, aiming to aggregate a unified outcome that closely aligns with the desired output. The averaging operation in this loss smooths out variances and biases from individual prompts, thus improving the overall prediction accuracy and reliability. 
Formally, this loss is formulated as
\begin{align}
p_{fused} &= \frac{1}{K}\sum_{i=1}^{K}\text{LLM}(\text{concat}(sp_i, c_n)) \nonumber\\
\mathcal{L}_{fusion} &= \mathrm{NLL}(p_{fused}, target_n).
\end{align}
By utilizing the collective strengths of diverse prompts, this loss enhances the model’s ability to generate context-appropriate responses.
%where the $p_{fused}$ is the aggregated results from $K$ soft prompts. As the loss computes the negative log-likelihood loss between the aggregated output and the ground truth.

\subsection{Overall Objective Function}
The SPT framework hinges on the harmonious integration of the aforementioned loss functions, where each addresses a distinct aspect. The soft prompt loss (i.e., $\mathcal{L}^{LLM}$) ensures the LLM fidelity, the prompt selection loss (i.e., $\mathcal{L}_{selection}$) aligns the retriever's similarity assessment with the LLM's output, the context-prompt contrastive loss (i.e., $\mathcal{L}_{con}$) promotes diversity in prompt selection, and the prompt fusion learning loss (i.e., $\mathcal{L}_{fusion}$) enhance the overall performance for all the soft prompts. The overall objective of the SPT method is to minimize a composite loss function that encapsulates these individual components. Formally, the overall objective function $\mathcal{L}_{Total}$ for the SPT framework is formulated as
\begin{align}
\mathcal{L}_{Total} =& \sum_{i=1}^{K} \mathcal{L}^{LLM}_i +\lambda_1 \sum_{\substack{i,j=1\atop i \neq j}}^{K} \mathcal{L}_{con}(s_{c_i}, s_{c_j}) \nonumber\\ 
&+\lambda_2\mathcal{L}_{selection}+\lambda_3 \mathcal{L}_{fusion},
\end{align}
where $\lambda_1$, $\lambda_2$, and $\lambda_3$ are hyperparameters that control the relative contribution of each loss component. In our experiments, we simply set $\lambda_1$, $\lambda_2$, and $\lambda_3$ to be $1$, which could achieve good performance. %The first term aggregates the individual negative log-likelihood losses for each soft prompt, fostering accurate and contextually relevant response generation. $\mathcal{L}_{selection}$, ensures that the retriever's similarity scores are in harmony with the LLM's performance evaluations. Lastly, the prompt-contrastive loss $\mathcal{L}_{con}$ is summed over all unique pairs of prompts, promoting diversity and avoiding redundancy in prompt selection.

By minimizing $\mathcal{L}_{Total}$ during training, the SPT framework effectively balances the fidelity to the LLM, the accuracy of the retriever, and the diversity in prompt selection, leading to an adaptive dialogue generation system.

\subsection{Inference}
During inference, the dense retriever selects the most appropriate soft prompt from the soft prompt group based on the given context. This selected prompt, along with the context, is then fed into the LLM to decode the final result. Formally, for a given context $C$, soft prompt group $SP$, and dense retriever $Ret$, the inference process proceeds as
\begin{equation}
\begin{aligned}
i^* &= \mathop{\arg\max}_{1\le i\le K} Ret(C, SP),\\
pred &= \mathrm{LLM}(\text{concat}(sp_{i^*}, C)),
\end{aligned}
\end{equation}
where $sp_{i^*}$ denotes the selected soft prompt with index $i^*$ and $pred$ denotes the response generated by the LLM.

\begin{table*}[t]
\centering
\resizebox{\textwidth}{!}{%
\begin{tabular}{llllllllllll}
\hline
Model & F1 & BLEU & ROUGE-1 & ROUGE-2 & ROUGE-L & BERT$_{F1}$ & BERT$_{P}$ & BERT$_{R}$ & DIST-1 & DIST-2 & AVG${_\uparrow}$ \\ \hline
OPT-125M-PT & 10.79 & 1.61 & 14.36 & 2.67 & 13.25 & 53.15 & 53.90 & 52.91 & 3.94 & 13.67 & - \\
OPT-125M-SPT & 11.06 & 2.22 & 16.45 & 3.60 & 15.42 & 54.86 & 56.23 & 53.91 & 4.87 & 17.38 & 16.60\% \\ \hline
OPT-1.3B-PT & 8.16 & 1.82 & 11.48 & 2.22 & 10.29 & 55.31 & 57.12 & 53.93 & 4.87 & 17.19 & - \\
OPT-1.3B-SPT & 9.94 & 2.66 & 13.74 & 3.24 & 12.38 & 56.34 & 58.08 & 54.99 & 4.93 & 17.76 & 16.43\% \\ \hline
OPT-2.7B-PT & 8.67 & 1.77 & 11.84 & 2.30 & 10.61 & 56.25 & 58.48 & 54.49 & 5.18 & 18.61 & - \\
OPT-2.7B-SPT & 12.23 & \textbf{3.11} & 16.97 & 4.37 & \textbf{15.61} & \textbf{57.96} & \textbf{59.92} & 56.45 & \textbf{5.84} & 20.76 & 33.04\% \\ \hline
Llama2-7B-PT & 17.12 & 1.99 & 15.74 & 4.07 & 13.72 & 52.30 & 48.57 & 57.11 & 2.80 & 12.91 & - \\
Llama2-7B-SPT & \textbf{17.49} & 2.80 & \textbf{17.02} & \textbf{4.48} & 15.24 & 54.66 & 53.02 & \textbf{57.14} & 5.69 & \textbf{22.86} & 26.62\% \\ \hline
\end{tabular}%
}
\vskip -0.1in
\caption{Performance comparison of different LLMs across different model sizes. BERT$_{F1}$, BERT$_{P}$, and BERT$_{R}$ denote the BERT Score F1, Precision, and Recall. AVG$_\uparrow$ indicates the average improvement over the corresponding baseline method. Models appended with `-SPT' indicate the combination of the proposed SPT method with the corresponding LLM, while `-PT' indicates the conventional prompt tuning method. The best performance in each metric is in bold.
%and the second-best one is highlighted with underlines. 
}
\label{tab:main_result}
\vspace*{-5mm}

\end{table*}
\section{Experiments}
In this section, we empirically evaluate the proposed SPT model.% framework presents a comprehensive , comparing its performance with that of established baselines.

\subsection{Dataset}
We conduct experiments on the ConvAI2 dataset \cite{convai2}, a benchmark for personalized dialogue generation. It comprises 8,939 training and 1,000 validation multi-turn conversations sourced from crowdworkers. Each dialogue includes persona profiles, each of which has four to five sentences to describe the background of each speaker, and the conversational history between the two interlocutors. By following \cite{mutual_persona,lapdog}, our experiments employ a self-persona setting where only the speaking interlocutor's persona is revealed, maintaining the other's persona as obscured.

\subsection{Experimental Setup}
All experiments are based on two LLMs, including OPT \cite{opt} and Llama2 \cite{llama2} of different sizes, which serve as the foundation model for the proposed SPT method. We randomly initialize soft prompts using a standard Gaussian distribution. % with a mean equal to $0$, a variance equal to $1$\footnote{***what are the mean of covariance matrix of this distribution?}.
For OPT models, we set the soft prompt token length to $8$, and for the Llama2 model, we use a token length of $1$. The soft prompt group consists of $K=4$ candidates. Learning rates of different LLMs are recorded in Table~\ref{tab:hyper} in the Appendix. %\footnote{***it is better to list them in a table in the appendix. Added} 
The threshold $\Gamma$ in Eq.~\eqref{context_CL_loss} is set to $20$.%, determined after analyzing the training data.  
% The hyperparameters $\lambda_1$, $\lambda_2$, $\lambda_3$, and $\lambda_4$ are uniformly set to $1$, with a threshold $\Gamma$ of $20$, determined after analyzing the training data.

\subsection{Evaluation Metrics}
We evaluate our model using a suite of established metrics for persona-based dialogue generation, including Unigram F1, BLEU, ROUGE, BERT Score, and textual unigram/bigram distinctness (denoted by DIST-1 and DIST-2). %\footnote{***complete it:Added}. 
Unigram F1 measures the harmonic mean of precision and recall at the token level. BLEU \cite{bleu} and ROUGE \cite{rouge} evaluate the overlap of $n$-grams between the generated text and target reference. BERT score \cite{bertscore}, using the \emph{deberta-xlarge-mnli} model\footnote{\url{https://github.com/Tiiiger/bert_score}} as recommended for its improved performance over \emph{roberta-large}, captures the semantic similarity of text pairs. Unigram and bigram distinctness (denoted by DIST-1 and DIST-2) gauge the diversity of the generated text, where DIST$_{AVG}$ denotes the average of DIST-1 and DIST-2.

\subsection{Results}
Table~\ref{tab:main_result} illustrates that the proposed SPT consistently outperforms the baseline models across various metrics. Notably, the OPT-2.7B-SPT and Llama2-7B-SPT models exhibit significant performance improvements (i.e., 33.04\% and 26.26\%, respectively). Those improvements affirm the effectiveness of the proposed SPT method in fostering more diverse and personalized responses.

For baseline models, we can see that there exists a common trade-off between linguistic quality and diversity. Specifically, the Llama2-7B model scores 17.12 in F1 and 1.99 in BLEU, but its diversity seems not so good %saw only modest gains in diversity 
(i.e., 2.80 in DIST-1 and 12.91 in DIST-2). This is in contrast to the OPT-125M model, which has lower linguistic scores (i.e., 10.79 in F1 and 1.61 in BLEU) but higher distinctness (i.e., 3.94 in DIST-1 and 13.67 in DIST-2). Different from those models, the proposed SPT method significantly enhances both diversity and  linguistic quality, %alongside notable improvements in linguistic and semantic metrics, including F1, BLEU, and BERT Score, 
thereby avoiding the common compromise between linguistic enhancement and diversity.

\section{Ablation Studies}

In this section, we conduct ablation studies for the proposed SPT method.

\begin{table}[]
\centering
\resizebox{\linewidth}{!}{%
\begin{tabular}{llllll}
\hline
Model & F1 & BLEU & ROUGE-L & BERT$_{F1}$ & DIST$_{AVG}$ \\ \hline
Llama-7B-SPT & \textbf{17.49} & \textbf{2.80} & \textbf{15.24} & \textbf{54.66} & 14.27 \\
w/o CL & 15.95 & 2.00 & 13.17 & 52.80 & 14.23 \\
w/o FUSION & 16.02 & 1.90 & 13.24 & 52.89 & \textbf{14.69} \\
w/o SL & 16.39 & 1.93 & 13.71 & 53.75 & 13.06 \\ \hline
\end{tabular}%
}
\vskip -0.1in
\caption{The ablation study on the training losses. `w/o CL', `w/o FUSION', and `w/o SL' denote no context-prompt contrastive loss, no prompt fusion learning loss, and no prompt selection loss, respectively.}
\label{tab:strategy}
\vspace*{-5mm}

\end{table}

\subsection{Training Losses}

Table~\ref{tab:strategy} reveals the impact of different training losses on performance. Omitting the prompt fusion loss slightly increases the prediction diversity in terms of DIST$_{AVG}$ but reduces the overall performance in terms of F1, BLEU, ROUGE, and BERT Score. One possible reason is that the prompt fusion loss contributes to the linguistic quality at the cost of the diversity.  %where the other two losses mainly contribute to the diversity at the cost of linguistic metrics.\footnote{***any explanation?: Added} 
Excluding the context-prompt contrastive loss leads to a decline in all the evaluated metrics, which shows the effectiveness of the context-prompt contrastive loss. 
The absence of the prompt selection loss significantly affects the prediction diversity, causing the model to favor a single soft prompt, akin to utilizing a single prompt. The above results underscore the importance of each loss in enhancing the model performance and response diversity.

\begin{figure}[t]
    \centering
    \includegraphics[width=\linewidth]{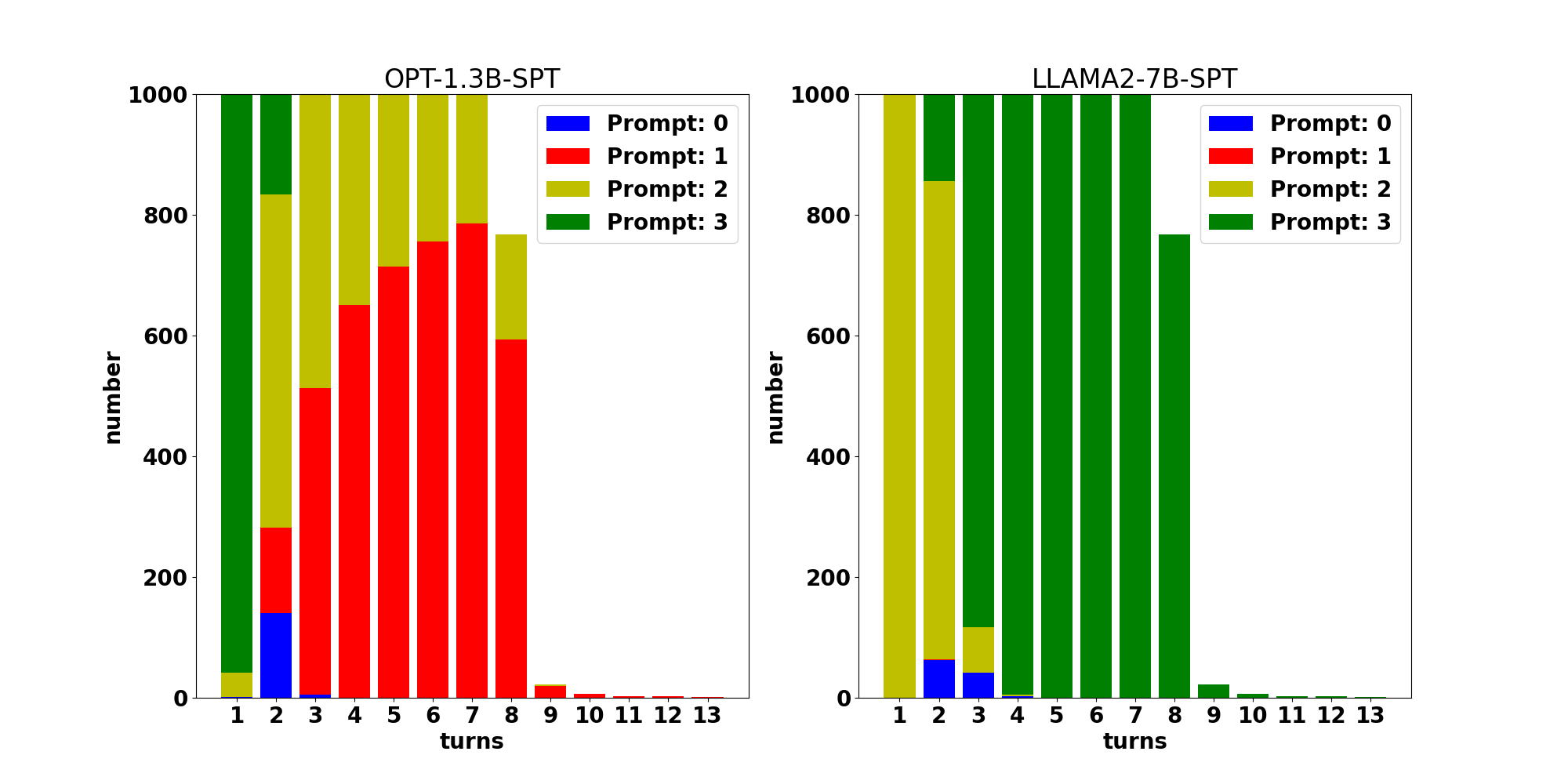}
    \vskip -0.1in
    \caption{Analysis of the usage of each soft prompt cross the conversational process, where the horizontal axis represents the index of the conversational turn and the vertical axis denotes the times that each soft prompt is chosen.
    %number of the selected index.
    }
\label{fig:abl_turns}
    \vspace*{-5mm}
\end{figure}

\begin{figure}[t]
    \centering
    \includegraphics[width=\linewidth]{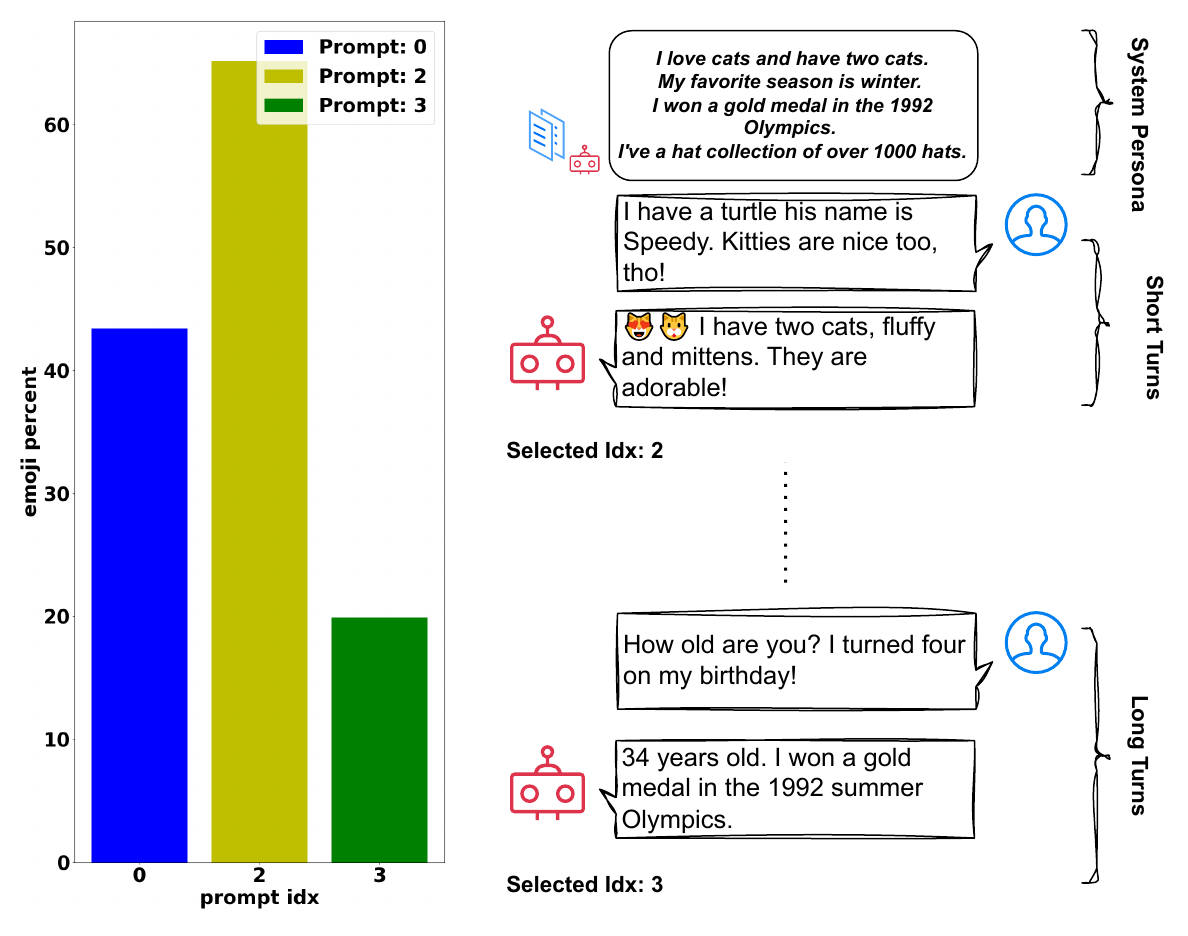}
    \vskip -0.1in
    \caption{The varied response styles of the Llama2-7B-SPT model, highlighting its tendency to incorporate emojis into responses during initial conversational turns.}
    \label{fig:abl_emoji}
    \vspace*{-5mm}
\end{figure}

\subsection{Prompt Usage in Varied Contexts}
%Our examination of the decoded text across different models revealed distinct patterns in prompt usage based on the length of the dialogue context.

To see the prompt usage during the conversational process, we plot in Figure~\ref{fig:abl_turns} the times each soft prompt is chosen during the entire conversation. 
According to Figure~\ref{fig:abl_turns}, we can see that in the OPT-1.3B-SPT model, prompt $sp_3$ is predominantly utilized for the initial stage in the conversation, $sp_2$ for the middle stage of the conversation, and $sp_1$ for the later stage of the conversation. For the Llama2-7B-SPT model, we have similar observations, indicating that soft prompts have functionalities in different stages of the conversation.% systematic assignment of prompts to dialogue turns.

Moreover, Figure~\ref{fig:abl_emoji} explores the stylistic aspects of responses generated by different prompts, i.e., emojis in the generated responses. In the Llama2-7B-SPT model, $sp_2$, which is often used in the initial stage of the conversations, tends to generate emojis in the generated response.  Differently, $sp_3$, often used in the late stage of the conversation, tends to generate few emoji in decoded responses. This phenomenon suggests a strategic use of emojis at different stages of the conversation.

% \usepackage{graphicx}
\begin{table}[]
\centering
\resizebox{\linewidth}{!}{%
\begin{tabular}{llllll}
\hline
$K$ & F1 & BLEU & ROUGE-L & BERT$_{F1}$ & DIST$_{AVG}$ \\ \hline
1 & 17.76 & 1.76 & 15.21 & 54.86 & 14.15 \\
2 & 17.71 & 2.55 & \textbf{15.63} & \textbf{55.52} & 14.29 \\
3 & 17.34 & 2.45 & 15.09 & 55.31 & \textbf{15.23} \\
4 & 17.49 & \textbf{2.80} & 15.24 & 54.66 & 14.27 \\
5 & 16.07 & 2.42 & 12.88 & 47.12 & 13.99 \\
6 & 17.46 & 2.21 & 14.94 & 54.43 & 15.12 \\
7 & \textbf{17.76} & 2.42 & 15.42 & 54.96 & 13.51 \\
8 & 17.48 & 2.32 & 15.29 & 54.87 & 13.89 \\ \hline
\end{tabular}%
}
\vskip -0.1in
\caption{The effect of the size of the soft prompt group (i.e., $K$) to the performance of Llama2-7B-SPT. % where DIST$_{AVG}$ denotes the average of the DIST-1 and DIST-2.
}
\label{tab:diff_k}
\vspace*{-5mm}

\end{table}

\subsection{Number of Soft Prompt Candidates}
Table~\ref{tab:diff_k} shows the effect of the number of soft prompts (i.e., $K$) to the model performance in terms of different metrics. %with optimal $K$ values differing for F1, BLEU, ROUGE-L, BERT$_{F1}$, and DIST$_{AVG}$. 
Though the best performance occurs at different $K$'s for different performance metrics, the best performance for different metrics usually occurs when $K\leq 4$, which is likely due to the sizes of both the CONVAI2 dataset and the LLM used. Hence, in all the experiments, $K$ is set to be 4 by default.%This emphasizes the importance of choosing the right number of soft prompts to strike a balance between accuracy and response diversity in the prompt fusion learning framework.

% \usepackage{graphicx}

\subsection{Comparison to Longer Prompt Tuning}
As shown in Table~\ref{tab:abl}, the SPT method with four single-token soft prompts outperforms the four-token prompt tuning method, highlighting effectiveness of the proposed SPT method. Moreover, SPT excels the eight-token prompt tuning method in terms of BLEU, ROUGE, and DIST$_{AVG}$, showing its effectiveness despite fewer trainable parameters.

% \subsection{Closest Text to Each Soft Prompt}
\subsection{Comparison to LoRA}
As LoRA \cite{lora} is another type of parameter-efficient finetuning method and has shown to be effective to utilize LLMs for different applications, we compare the proposed SPT method with it based on the Llama2-7B model under the condition that they have comparable numbers of trainable parameters. 
%To benchmark our approach and soft prompt tuning against LoRA , we applied LoRA tuning to the Llama2-7B model under similar parametric conditions. 
As shown in Table~\ref{tab:abl}, LoRA exhibits improvements in the BLEU score and DIST$_{AVG}$ but has lower ROUGE-L, BERT$_{F1}$, and F1 scores compared with the four-token prompt tuning method. Moreover, the proposed SPT method surpasses LoRA across all the evaluation metrics, highlighting its superior performance and affirming its effectiveness under the condition of comparable numbers of trainable parameters.

\subsection{Comparison to In-Context Learning}
To compare the performance with In-Context Learning (ICL) on LLMs, we compare the SPT method with the zero-shot GPT-3.5 turbo with instructions. According to results shown in Table~\ref{tab:abl}, we can see that ICL gains a higher diversity score (i.e., DIST$_{AVG}$) but lower scores in terms of other metrics. This implies that simply prompting a more powerful LLM without proper tuning is hard to gain comparable performance to tuning methods.
% \usepackage{graphicx}
\begin{table}[]
\centering
\resizebox{\linewidth}{!}{%
\begin{tabular}{llllll}
\hline
Model & F1 & BLEU & ROUGE-L & BERT$_{F1}$ & DIST$_{AVG}$ \\ \hline
Llama2-7B-SPT & 17.49 & \textbf{2.80} & \textbf{15.24} & 54.66 & 14.27 \\
Llama2-7B-4-PT-TOKEN & 16.47 & 1.78 & 13.64 & 52.65 & 9.52 \\
Llama2-7B-8-PT-TOKEN & \textbf{17.64} & 2.13 & 14.69 & \textbf{55.85} & 13.33 \\ \hline
Llama2-7B-LoRA & 15.61 & 2.20 & 11.66 & 47.46 & 10.21 \\ \hline
GPT-3.5-ICL & 6.78 & 0.77 & 0.09 & 47.96 & \textbf{23.24} \\ \hline
\end{tabular}%
}
\vskip -0.1in
\caption{Performance comparison across varying prompt token lengths as well as LoRA and In-Context Learning on GPT-3.5 Turbo. `-SPT' denotes the proposed SPT model with a single token length per prompt, while Llama2-7B-4-PT-TOKEN and Llama2-7B-8-PT-TOKEN have token lengths of 4 and 8, respectively.}
\label{tab:abl}
% \vspace{-3mm}
\end{table}

% \usepackage{graphicx}
\begin{table}[]
\centering
\resizebox{\linewidth}{!}{%
\begin{tabular}{llllll}
\hline
Model & BLEU & BLEU-1 & BLEU-2 & BLEU-3 & BLEU-4 \\ \hline
Llama2-7B-PT & 8.79 & 43.42 & 20.44 & 13.51 & 10.06 \\
Llama2-7B-SPT & 2.07 & 41.99 & 16.62 & 10.48 & 6.95 \\ \hline
\end{tabular}%
}
\vskip -0.1in
\caption{Comparison of text overlapping between the prediction of different models and the persona.}
\label{tab:bleu_to_persona}
\vspace*{-5mm}
\end{table}

\subsection{Text Overlap Between Prediction and Persona}

Table~\ref{tab:bleu_to_persona} presents BLEU scores between the model's predictions and the system's persona descriptions for different models. We can see that the prompt tuning method exhibit larger text overlap with the system's persona, often leading to repetitive responses aligned with the persona. In contrast, the proposed SPT method has lower linguistic similarities to the persona, which results in more diverse and effective responses. This suggests that the proposed SPT method effectively balances the persona consistency and response diversity, avoiding the pitfalls of over-repetition.

\section{Conclusion}
In this paper, we introduce SPT, a strategic approach for personalized dialogue generation through selective prompt tuning. By jointly training a soft prompt group and a dense retriever, SPT adeptly navigates various conversational scenarios automatically, enriching response diversity while improving both linguistic and neural-based metrics. Experiments on the CONVAI2 dataset highlights the capacity of SPT to identify intrinsic conversational settings, showing its effectiveness in generating contextually appropriate dialogues. %This advancement opens new pathways for conversational AI research, promising more engaging and personalized interaction models.

\section*{Acknowledgements}

This work is supported by NSFC general grant 62076118 and Shenzhen fundamental research program JCYJ20210324105000003.

\section*{Limitations}
This paper has introduced the selective prompt tuning in personalized dialogue generation. Through diverse prompting, the LLMs can generate more diverse and engaged responses when compared with single prompt tuning. However, despite the context-prompt contrastive mechanism and prompt selection loss, there is still a risk for the retriever to fall into a narrow selection of soft prompts (e.g., given $K=4$ in Llama2-7B, there is still one soft prompt that is selected only once during inference). This limitation may caused by a larger $K$ used, making the determination of $K$ important. %insufficient conversation discovery of the SPT. 
Meanwhile, in the context-prompt contrastive loss, simply using BLEU to measure text similarity may not be sufficient to distinguish the difference between two dialogues, which could be enhanced by neural metrics powered by LLMs that could distinguish texts from both semantic and linguistic perspectives. Additionally, in the decoded text of Llama2-7B, the existence of emoji is not designed in the PersonaChat dataset, which is worth further investigation. 

\section*{Ethic Statement}
This research confines the use of personal data to fictional persona profiles in the CONVAI2 dataset, avoiding the handling or storage of real personal data. All the soft prompts within the SPT are vector-based parameters without directly encoding or representing any individual's personal information. When applying to real-world applications, it is vital to prioritize data privacy, ensuring that personal information for personalized dialogues is ethically sourced and used with informed consent. 

\bibliography{acl_latex}

\clearpage

\appendix

\section{Appendix}
\algrenewcommand\algorithmicrequire{\textbf{Input:}}
\algrenewcommand\algorithmicensure{\textbf{Output:}}

\begin{algorithm*}[!t]
\caption{SPT Training}
\label{alg:training}
\begin{algorithmic}[1]
\Require Input context $C=\{c_1,..,c_N\}$, Input context batch $C_{batch}=\{c_i,..,c_{i+batch size}\} \subset C$, ground truth response $Y=\{y_1, ..., y_N\}$, a soft prompt group $SP=\{sp_1,.., sp_K\}$, a dense retriever $Ret$, textual similarity threshold $\Gamma$, a text similarity metric $M$, and a large language model $LLM$
\Ensure{A tuned soft prompt group $SP$ and a tuned dense retriever $Ret$}
\For {$C_{batch}$ in $C$}
\State Initialize batch soft prompt loss $\mathcal{L}_{batch}^{LLM}=0$, batch prompt selection loss $\mathcal{L}_{selection}^{batch}=0$, 
\State Initialize batch prompt fusion loss $\mathcal{L}_{fusion}^{batch}=0$, batch context-prompt contrastive loss $\mathcal{L}_{con}^{batch}=0$

\For{Input Context $c_n$ in $C_{batch}$}
\State Compute one soft prompt $\mathcal{L}_i^{LLM}=NLLLoss(concat(sp_i, c_n),y_n)$
\State Obtain $K$ soft prompt loss $\mathcal{L}^{LLM}=\{\mathcal{L}_1^{LLM},..., \mathcal{L}_K^{LLM}\}$ with above computation
\State Normalized negative soft prompt loss $\mathcal{L}^{LLM}_{normed}= \text{Softmax}(-\mathcal{L}^{LLM}/\tau_g)$ for retriever update
\State Compute retriever score between context $c_n$ and soft prompt $sp_i$ as $s_{c_n, sp_i}=Ret(sp_i, c_n)$
\State Obtain $K$ retriever scores by $s_{c_n, SP}=\{s_{c_n, sp_1},..., s_{c_n, sp_K}\}$
\State Compute prompt selection loss using KL Divergence by $\mathcal{L}_{selection} = \text{KL}(s_{c_n, SP},\mathcal{L}^{LLM}_{normed})$
\State Aggregate $K$ predictions from LLM given $c_n$ and $SP$ as $p_{fused}$
\State Compute prompt fusion loss as $\mathcal{L}_{fusion} = \mathrm{NLL}(p_{fused}, y_n)$ 
\State Sum soft prompt loss, prompt selection loss, and prompt fusion loss to their batch opponents
\State $\mathcal{L}_{batch}^{LLM}=\mathcal{L}_{batch}^{LLM}+\mathcal{L}^{LLM}$, $\mathcal{L}_{selection}^{batch}=\mathcal{L}_{selection}^{batch}+\mathcal{L}_{selection}$, $\mathcal{L}_{fusion}^{batch}=\mathcal{L}_{fusion}^{batch}+\mathcal{L}_{fusion}$
\EndFor
\For{Input Context $c_i,c_j$ in $C_{batch}$}
\State Compute textual similarity $T=M(c_i,c_j)$
\State Compute retriever score for $c_i, c_j$ as $s_{c_i,SP},s_{c_j,SP}$
\State Compute context-prompt contrastive loss:
\If $T>\Gamma$
\State $\mathcal{L}_{con}=1-cos(s_{c_i,SP},s_{c_j,SP})$
\Else
\State $\mathcal{L}_{con}=\max(0, \cos(s_{c_i,SP},s_{c_j,SP}))$
\EndIf
\State Sum context-prompt contrastive loss to batch context-prompt contrastive loss 
\State $\mathcal{L}_{con}^{batch}=\mathcal{L}_{con}^{batch}+\mathcal{L}_{con}$
\EndFor
\State Sum all objective together: $\mathcal{L}_{Total}=\mathcal{L}_{batch}^{LLM}+\mathcal{L}_{selection}^{batch}+\mathcal{L}_{fusion}^{batch}+\mathcal{L}_{con}^{batch}$
\State Update soft prompts and retriever via back-propagation with $\mathcal{L}_{Total}$
\EndFor
\end{algorithmic}
\end{algorithm*}

\subsection{Complete Training Procedure}
The full training procedure is described at Algorithm~\ref{alg:training}.

% \usepackage{graphicx}

\subsection{Detailed Settings for SPT Training}
\begin{table}[h]
\centering
\begin{tabular}{ll}
\hline
\multicolumn{2}{c}{Shared   Parameters} \\ \hline
HyperParameter & Value \\ \hline
K & 4 \\
Optimizer & Adam \\
$\tau_g$ & 1 \\
$\lambda_1$ & 1 \\
$\lambda_2$ & 1 \\
$\lambda_3$ & 1 \\
$\lambda_4$ & 1 \\ \hline
\multicolumn{2}{c}{Llama2-7B-SPT} \\ \hline
Prompt Length & 1 \\
Learning Rate & 0.01 \\ \hline
\multicolumn{2}{c}{OPT-2.7B} \\ \hline
Prompt Length & 8 \\
Learning Rate & 0.001 \\ \hline
\multicolumn{2}{c}{OPT-1.3B} \\ \hline
Prompt Length & 8 \\
Learning Rate & 0.01 \\ \hline
\multicolumn{2}{c}{OPT-125M} \\ \hline
Prompt Length & 8 \\
Learning Rate & 0.01 \\ \hline
\end{tabular}%

\caption{The hyper-parameters for the SPT training.}
\label{tab:hyper}
\end{table}

Table~\ref{tab:hyper} lists the detailed hyper-parameters for training SPT. The share parameters are used for all model training. Meanwhile, the Llama2-7B-SPT, OPT-2.7B, OPT-1.3B, and OPT-125M indicate the specific hyper-parameters used in the specific model training. We trained the SPT models on eight Tesla-V100 32GB GPUs. For each SPT model except OPT-125M-SPT, we train one epoch and then do the evaluation. For OPT-125M-SPT, we train for 15 epochs until it converges.

\subsection{Details for Ablation Study}
Table~\ref{tab:abl_detail} details our ablation study's findings. Selective Prompt Tuning (SPT) with four one-token soft prompts demonstrates superior performance over both the traditional four-token and eight-token soft prompt tuning approaches, highlighting our method's effectiveness. In a comparative analysis with LoRA under a similar parameter setup, SPT outperforms in all evaluated metrics, reinforcing its efficiency. Furthermore, compared to GPT-3.5 Turbo's In-Context Learning (ICL), SPT shows significant improvements in F1 and BLEU scores, indicating challenges with ICL's alignment to target responses despite its higher diversity in textual outputs.

% The details for the ablation study are shown in Table~\ref{tab:abl_detail}. As we can see, the SPT with four one-token soft prompts outperforms one four-token soft prompt tuning. Meanwhile, it also gains advantages when compared to the eight-token soft prompt-tuning opponent. Demonstrating the effectiveness. Simultaneously, we compared our approach with LoRA with a similar parametric setting. Compared to four-token prompt tuning, LoRA obtained improvements in BLEU and distinctness, but lower ROUGE and BERT scores. When compared to SPT, SPT outperforms LoRA in all metrics, which affirms the effectiveness under a similar parametric setting. As compared to the in-context learning (ICL) from GPT-3.5 Turbo, the low F1 and BLEU indicates that the ICL approach is hard to align with the target response through simply textual prompting, regardless of its high textual diversity. 

% \footnote{***did not see the proposed SPT method.: In the first row.}\footnote{***add some analysis}

% \usepackage{graphicx}
\begin{table}[!htbp]
\centering
\resizebox{\linewidth}{!}{%
\begin{tabular}{llll}
\hline
 & Persona Consistency & Dialogue Consistency & Engageness \\ \hline
Llama2-7B-SPT & \textbf{1.89} & \textbf{1.29} & \textbf{1.34} \\
Llama2-7B-PT & 1.33 & 1.13 & 1.29 \\ \hline
\end{tabular}%
}
\caption{Human evaluation over Llama2-7B-SPT and Llama2-7B-PT.}
\label{tab:human}
\end{table}

\subsection{Human Evaluation}
We conducted human evaluation on three metrics, persona consistency, context consistency, and engagingness. Each metric is ranked for three scores: 0, 1, 2. For persona consistency, 0 means contradicts the persona, 1 means not relevant to the persona, and 2 means consistent to the persona. For context consistency, 0 means contradicts previous dialogue history, 1 means not relevant to the previous dialogue, and 2 means consistent to the previous dialogue. For engagingness, 0 means a boring response, 1 means a safe but bland response, and 2 means an interesting response. We randomly sampled 100 responses from Llama2-7B-SPT and Llama2-7B-PT. The results are displayed in Table~\ref{tab:human}. Our proposed SPT outperforms PT over all three metrics, indicating the effectiveness of our approach in both three perspectives.
% \usepackage{graphicx}
\begin{table*}[h]
\centering
\resizebox{\textwidth}{!}{%
\begin{tabular}{lllllllllll}
\hline
Model & F1 & BLEU & ROUGE-1 & ROUGE-2 & ROUGE-L & BERT$_{F1}$ & BERT$_{P}$ & BERT$_{R}$ & DIST-1 & DIST-2 \\ \hline
Llama-7B-SPT & 17.49 & 2.80 & \textbf{17.02} & \textbf{4.48} & \textbf{15.24} & 54.66 & 53.02 & \textbf{57.14} & \textbf{5.69} & \textbf{22.86} \\
Llama2-7B-4-PT-TOKEN & 16.47 & 1.78 & 15.74 & 3.74 & 13.64 & 52.65 & 49.09 & 57.18 & 3.35 & 15.70 \\
Llama2-7B-8-PT-TOKEN & \textbf{17.64} & 2.13 & 16.49 & 4.01 & 14.69 & \textbf{55.85} & \textbf{54.98} & 57.34 & 4.75 & 21.91 \\ \hline
LoRA & 15.61 & 2.20 & 13.09 & 2.95 & 11.66 & 47.46 & 47.78 & 47.48 & 3.35 & 17.08 \\ \hline
GPT-3.5-ICL & 6.78 & 0.77 & 0.00 & 0.00 & 0.09 & 47.96 & 46.73 & 49.77 & \textbf{8.03} & \textbf{38.45} \\ \hline
\end{tabular}%
}
\caption{Detailed results for the ablation study.}
\label{tab:abl_detail}
\end{table*}

\subsection{Experimental Results on Larger Dataset}
To further evaluate the efficiency and scalability of the SPT framework. We conducted additional experiments on the DailyDialog dataset, a more extensive and complex dialogue dataset than PersonaChat. Notably, the DailyDialog dataset lacks explicit persona descriptions in its entries, presenting a unique challenge for personalization techniques. The results of the DailyDalog are shown as Table~\ref{tab:dailydialog}.

\paragraph{Result Analysis}: The experimental setup involved executing four separate runs using both soft prompt tuning (PT) and SPT strategies on the DailyDialog dataset. The empirical evidence clearly demonstrates the superiority of the SPT framework over the conventional PT approach across all evaluated metrics. Specifically, the SPT method exhibits significant performance improvements, showcasing its adaptability and effectiveness in handling more complex and extensive datasets. The evaluation metrics are summarized in the table below, where we observe notable enhancements in key areas such as F1 score, BLEU, ROUGE, and BERT-based metrics, underlining SPT's potential applicability across diverse conversational tasks.
% \usepackage{graphicx}
\begin{table*}[h]
\centering
\resizebox{\textwidth}{!}{%
\begin{tabular}{lllllllllll}
\hline
Model & F1 & BLEU & ROUGE1 & ROUGE2 & ROUGEL & BERT F1 & BERT Precision & BERT Recall & DIST-1 & DIST-2 \\ \hline
Llama2-7B-PT-LR=0.001 & 18.03 & 0.18 & 15.66 & 4.41 & 14.13 & 55.90 & 55.97 & 56.99 & 7.71 & 35.40 \\
Llama2-7B-PT-LR=0.01 & 17.06 & 0.21 & 14.46 & 4.09 & 13.00 & 54.58 & 54.71 & 55.60 & 7.41 & 34.83 \\
Llama2-7B-SPT-LR=0.001 & \textbf{18.38} & \textbf{0.31} & \textbf{15.87} & \textbf{4.49} & \textbf{14.37} & \textbf{56.95} & \textbf{57.21} & \textbf{57.73} & \textbf{7.97} & \textbf{36.89} \\
Llama2-7B-SPT-LR=0.01 & 17.40 & 0.08 & 15.04 & 4.57 & 13.68 & 53.72 & 53.73 & 55.16 & 7.53 & 34.68 \\ \hline
\end{tabular}%
}
\caption{Detailed results for DailyDialog.}
\label{tab:dailydialog}
\end{table*}

\subsection{Comparison to RAG (Retrieval Augmented Generation)}

\paragraph{Conceptual Differences:} RAG and SPT fundamentally differ in their approaches. RAG enhances inputs by incorporating external information from a database, focusing on the value of external data. In contrast, SPT focuses on selecting the optimal soft prompt based on given context input. While they operate differently, they aren't inherently conflicting and could be seen as complementary since SPT can treat the retrieval-augmented input as context as a whole. SPT has the potential to integrate RAG's enriched inputs comprehensively. The exploration of combining RAG and SPT falls beyond the scope of this work and is reserved for future research.

\paragraph{RAG Experimentation:} We experimented with the RAG framework under the Llama2-7B model to compare SPT with RAG. We observed that the choice of $K$ (number of retrieval contents) is crucial due to the RAG's reliance on the training set for retrieval. A large $K$ value can lead to the concatenated content overwhelming the context window size, thus significantly increasing computational resource demands.

\paragraph{Efficient Training Setup for RAG:} For efficiency, we set $K=1$ for our RAG experiment, focusing on retrieving the most semantically similar dialogue to augment the current context. The retriever used is the Contriever from Facebook, which is known for its ability to retrieve highly relevant content based on textual semantics. This setup allowed us to directly compare the efficiency and scalability of RAG and SPT under similar computational constraints.

\paragraph{Comparative Results:} The training time for an epoch under the RAG setup was approximately 14 hours, compared to 7 hours for SPT. This underscores SPT's efficiency and scalability, especially in resource-constrained environments. Detailed results are displayed in the table~\ref{tab:rag}. In terms of performance, SPT outperformed RAG in nearly all the metrics. This shows that SPT is not only faster but can also produce better results. The only area where RAG did slightly better was in creating more diverse responses (DIST-1 and DIST-2 metrics). This comparison shows that SPT is more efficient and often more effective than RAG. However, these two approaches do not necessarily contradict each other. Instead, combining these two methods could lead to even better performance. We might create more accurate and engaging dialogues by using RAG to get the proper context and SPT to fine-tune the response. This approach has a lot of potential for improving conversational AI systems.

% \usepackage{graphicx}
\begin{table*}[h]
\centering
\resizebox{\textwidth}{!}{%
\begin{tabular}{lllllllllll}
\hline
Model & F1 & BLEU & ROUGE-1 & ROUGE-2 & ROUGE-L & BERT-F1 & BERT-Precision & BERT-Recall & DIST-1 & DIST-2 \\\hline
RAG-Contriever-LR=0.01 & 14.93 & 2.18 & 9.36 & 2.03 & 8.55 & 45.01 & 45.87 & 44.72 & 4.95 & 24.47 \\
RAG-Contriever-LR=0.001 & 12.66 & 2.59 & 9.66 & 2.48 & 8.89 & 40.72 & 41.60 & 40.35 & 5.13 & 24.81 \\
RAG-Contriever-LR=0.0001 & 15.16 & 2.53 & 11.53 & 2.86 & 10.56 & 50.52 & 50.70 & 51.12 & \textbf{5.75} & \textbf{26.83} \\
Llama2-7B-SPT & \textbf{17.49} & \textbf{2.80} & \textbf{17.02} & \textbf{4.48} & \textbf{15.24} & \textbf{54.66} & \textbf{53.02} & \textbf{57.14} & 5.69 & 22.86\\
\hline
\end{tabular}%
}
\caption{The comparison between RAG and SPT.}
\label{tab:rag}
\end{table*}

\subsection{SPT Stability Experiment}
To evaluate the stability of the SPT, we further conducted additional experiments designed to test the system's resilience to disruptions. Specifically, we introduced Gaussian noises with the mean as 0 and the standard deviation as 1 to the similarity scores during inference to simulate the effect of inaccuracies in the soft prompt selection process. Additionally, we add a parameter $\alpha$ to control the strength of the noise. Formally, the disrupted selection score would become $score = score + \alpha * noise$. The objective of this experiment is to observe the stability of our retriever under less-than-ideal conditions. Detailed results of these experiments will be included in our revision.

\paragraph{Result Analysis}: The results presented in Table~\ref{tab:stable} demonstrate the impact of noise on retrieval performance. The introduction of mild noise (e.g., 0.01 to 0.1) results in negligible performance degradation, with some metrics showing slight improvements. However, as noise levels increase to 1.0, a deterioration in performance is observed despite a noticeable increase in DIST-2. This pattern suggests that while our SPT framework exhibits good stability to minor disturbances, its performance is adversely affected by severe interference.

% \usepackage{graphicx}
\begin{table*}[]
\centering
\resizebox{\textwidth}{!}{%
\begin{tabular}{lllllllllll}
\hline
Model & F1 & bleu & rouge1 & rouge2 & rougel & BERT f1 & BERT precision & BERT recall & dist-1 & dist-2 \\\hline
SPT (Noise=0) & \textbf{17.49} & 2.80 & \textbf{17.02} & 4.48 & \textbf{15.24} & 54.66 & 53.02 & \textbf{57.14} & \textbf{5.69} & 22.86 \\
Noise=0.01 & 17.42 & 2.97 & 16.93 & \textbf{4.49} & 15.18 & 54.70 & 53.45 & 56.71 & 5.42 & 22.55 \\
Noise=0.05 & 17.41 & 2.98 & 16.91 & 4.48 & 15.16 & 54.68 & 53.41 & 56.71 & 5.41 & 22.56 \\
Noise=0.1 & 17.42 & 2.93 & 16.91 & 4.44 & 15.17 & 54.75 & 53.46 & 56.78 & 5.46 & 22.86 \\
Noise=0.5 & 17.41 & \textbf{2.99} & 16.92 & \textbf{4.49} & 15.16 & \textbf{54.93} & \textbf{53.56} & 57.03 & 5.61 & 24.04 \\
Noise=1.0 & 17.43 & 2.76 & 16.72 & 4.32 & 15.03 & 54.82 & 53.38 & 57.00 & 5.56 & \textbf{24.15}\\\hline
\end{tabular}%
}
\caption{The experiment on the stability of the SPT.}
\label{tab:stable}
\end{table*}

\begin{table*}[h!]
\centering
\resizebox{\textwidth}{!}{%
\begin{tabular}{lllllllllll}
\hline
Model & F1 & BLEU & ROUGE-1 & ROUGE-2 & ROUGE-L & BERT f1 & BERT precision & BERT recall & DIST-1 & DIST-2 \\\hline
SPT (Noise=0) & 17.49 & 2.80 & 17.02 & 4.48 & 15.24 & 54.66 & 53.02 & 57.14 & 5.69 & 22.86 \\
Noise=0.001 & 17.69 & 2.75 & 17.11 & 4.46 & 15.38 & 54.94 & 53.41 & 57.19 & 5.54 & 23.83 \\
Noise=0.01~1.0 & NaN & NaN & NaN & NaN & NaN & NaN & NaN & NaN & NaN & NaN\\\hline
\end{tabular}%
}
\caption{The experiment on dense retriever stability.}
\label{tab:ret_stable}
\end{table*}

\subsection{Retriever Stability Experiment}
To evaluate the robustness of our dense passage retrieval system, we introduced Gaussian noise with standard deviation. Specifically, we apply noise with a varying strength $\alpha$, choosing from [0.001, 0.01, 0.1, 0.2, 0.3, 0.4, 0.5, 0.6, 0.7, 0.8, 0.9, 1.0], to the $L^{LLM}_{normed}$ loss during the retriever's training phase. Therefore, the disrupted $L^{LLM}_{normed}$ will become $L^{LLM}_{normed}+\alpha * noise$. This approach aimed to simulate potential disruptions in the soft prompt selection process, thereby testing the stability and resilience of our retriever under adversarial conditions.

\paragraph{Adversarial Noise Impact on Retriever Robustness:} The introduction of Gaussian noise served as a means to disturb the updating process of the retriever, allowing us to observe its behaviour and adaptability in the interference. Specifically, we add the noise the $\mathcal{L}^{LLM}_{normed}$ to make the KL Divergence update become noisy. The varying levels of noise strength were chosen to represent a wide spectrum of potential adversarial impacts, from mild to severe disruptions, i.e., [0.001, 0.01, 0.1, 0.2, 0.3, 0.4, 0.5, 0.6, 0.7, 0.8, 0.9, 1.0]

\paragraph{Results and Insights:} According to the Table~\ref{tab:ret_stable}, introducing the mildest level of noise (0.001) yielded improved performance across several key metrics, including F1, ROUGE-1, ROUGE-L, BERT Score, and DIST-2. This improvement suggests that slight perturbations may act as a beneficial regularizer within the training process, thereby enhancing performance. In contrast, levels of noise beyond the mildest introduced numerical instability (manifesting as overflow or underflow, particularly as we utilize fp16 for SPT training). This instability disrupts the training process, leading to outcomes marked as NaN (Not a Number).

% \usepackage{graphicx}

\subsection{Case Study}

\begin{figure*}[h!]
    \centering
    \includegraphics[width=\textwidth]{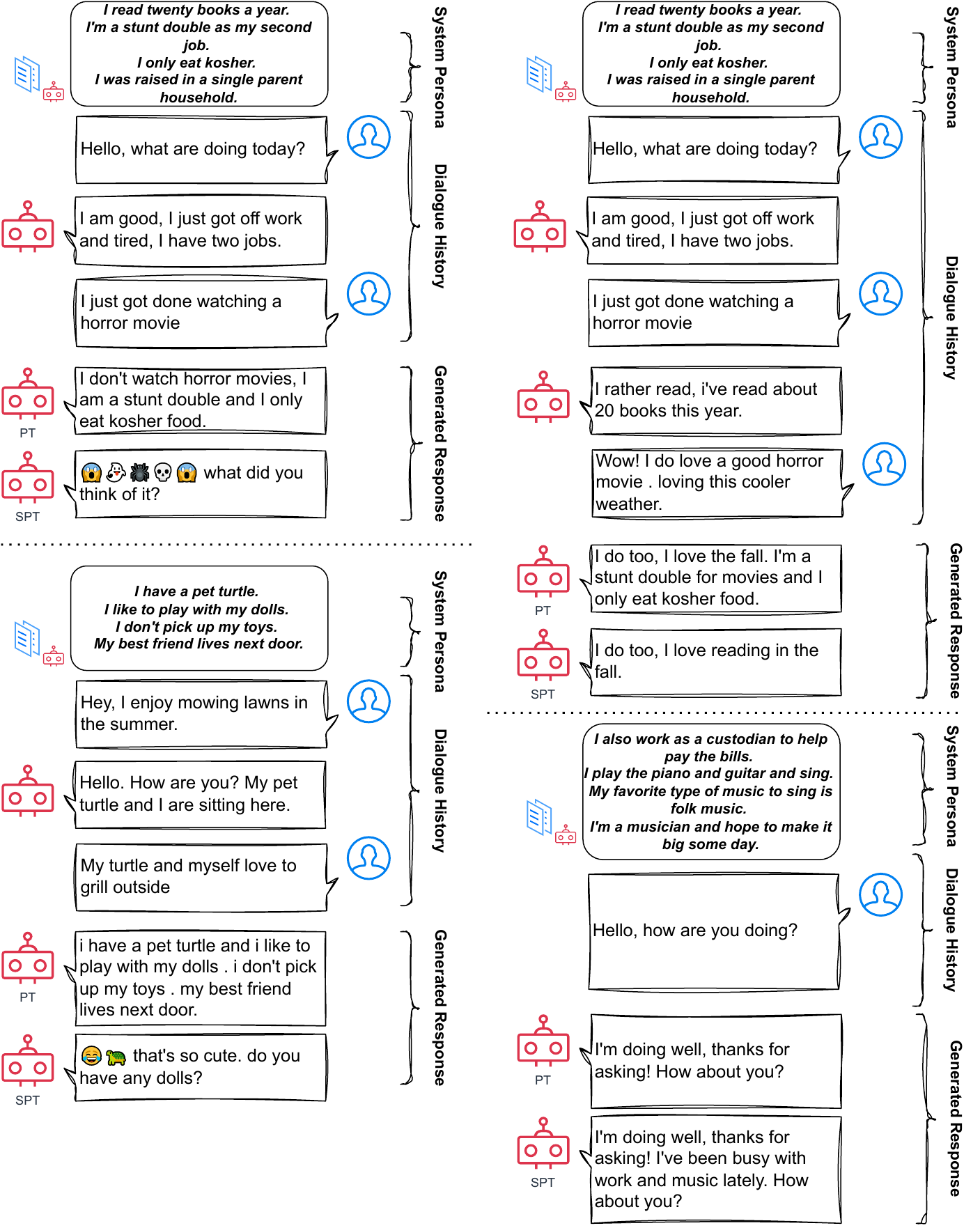}
    \caption{Four case studies, where PT denotes the prompt tuning method \cite{prompt-tuning}.}
    \label{fig:case_study}
\end{figure*}

Figure~\ref{fig:case_study} shows a comparison between SPT and a prompt-tuned model. SPT uniquely incorporates horror-related emojis in a conversation about horror movies, while the prompt-tuned model tends to repeat persona profile content. This trend continues in subsequent dialogues. In the last case, SPT adeptly weaves persona details into its responses, offering a more engaging and personalized conversational experience compared to the more generic replies of the prompt-tuned model.

\end{document}